\documentclass[10pt,twocolumn,letterpaper]{article}
\usepackage{CJKutf8}  
\usepackage[pagenumbers]{cvpr} 

\usepackage{amssymb} 
\usepackage{pifont}
\usepackage{times}
\usepackage{graphicx}
\usepackage{amsmath}
\usepackage{booktabs}          
\usepackage{multirow}          
\usepackage{xcolor}            
\usepackage{tabularx}
\usepackage{makecell}
\usepackage{adjustbox}
\usepackage{colortbl} 

\definecolor{yellow}{HTML}{F4FF7F} 

\definecolor{cvprblue}{rgb}{0.21,0.49,0.74}

\newif\ifshowcomments
\showcommentstrue

\ifshowcomments

\newcommand{\xwhu}[1]{{\color{blue}{[XW: #1]}}}

\usepackage[pagebackref,breaklinks,colorlinks,citecolor=cvprblue]{hyperref}

\def\paperID{261} 
\def\confName{CVPR}
\def\confYear{2025}

\title{EchoInk-R1: Exploring Audio-Visual Reasoning  \\ in Multimodal LLMs   via Reinforcement Learning}

\author{
Zhenghao Xing$^{1}$,
Xiaowei Hu$^{2,}$\thanks{Corresponding author (huxiaowei@pjlab.org.cn).},
Chi-Wing Fu$^{1}$, \\
Wenhai Wang$^{2}$,
Jifeng Dai$^{3,2}$, and
Pheng-Ann Heng$^{1}$ \vspace{1mm} \\
$^1$ The Chinese University of Hong Kong \\
$^2$ Shanghai Artificial Intelligence Laboratory
$^3$ Tsinghua University
\vspace{1mm} \\
\url{https://github.com/HarryHsing/EchoInk}
}

\begin{document}
\begin{CJK}{UTF8}{gbsn}  
\maketitle
\makeatletter
\renewcommand{\@makefnmark}{{\dag}}  
\makeatother

\begin{abstract}

Multimodal large language models (MLLMs) have advanced perception across text, vision, and audio, yet they often struggle with structured cross-modal reasoning, particularly when integrating audio and visual signals. We introduce \textbf{EchoInk-R1}, a reinforcement learning framework that enhances such reasoning in MLLMs. 
Built upon the Qwen2.5-Omni-7B foundation and optimized with Group Relative Policy Optimization (GRPO), EchoInk-R1 tackles multiple-choice question answering over synchronized audio-image pairs. To enable this, we curate \textbf{AVQA-R1-6K}, a dataset pairing such audio-image inputs with multiple-choice questions derived from OmniInstruct-v1.
EchoInk-R1-7B achieves 85.77\% accuracy on the validation set, outperforming the base model, which scores 80.53\%, using only 562 reinforcement learning steps.
Beyond accuracy, EchoInk-R1 demonstrates reflective reasoning by revisiting initial interpretations and refining responses when facing ambiguous multimodal inputs. These results suggest that lightweight reinforcement learning fine-tuning enhances cross-modal reasoning in MLLMs.
EchoInk-R1 is the first framework to unify audio, visual, and textual modalities for general open-world reasoning via reinforcement learning. Code and data are publicly released to facilitate further research.

\end{abstract}

\section{Introduction} 
\label{sec:intro}

Recent advances in multimodal large language models (MLLMs) have significantly improved the perception capabilities across text, vision, and audio modalities \cite{zhang2023video,cheng2024videollama,xu2025qwen2,guo2025aligned,xing2025echotraffic}.
These systems can interpret rich multimodal inputs and generate structured outputs, enabling applications such as open-domain question answering, multimedia retrieval, and interactive agents.
However, despite their perceptual competence, current MLLMs still lack robust multimodal reasoning abilities, often relying on shallow correlations rather than coherent, multi-step, and cross-modal inference, especially when integrating auditory, visual, and textual signals.

Meanwhile, recent works such as OpenAI’s o1 \cite{openai2024o1} and DeepSeek-R1 \cite{guo2025deepseek} demonstrate that reinforcement learning (RL) can significantly enhance reasoning abilities in large models.
By optimizing structured reasoning objectives, these methods enable large language models to perform complex step-by-step deductions.
However, existing RL-enhanced models mainly focus on text-only \cite{guo2025deepseek, hu2025open, yu2025dapo}, audio-language \cite{li2025reinforcementlearningoutperformssupervised, xie2025audioreasonerimprovingreasoningcapability, wen2025saristructuredaudioreasoning}, or vision–language reasoning \cite{feng2025video, zhou2025visualthinker, meng2025mmeureka}, with limited attention given to integrated audio-visual reasoning.
Currently, no prior work achieves open-world reasoning across audio, visual, and textual modalities within a unified reinforcement learning framework.

\begin{figure*}[tp]
    \centering
    \includegraphics[width=0.9\hsize]{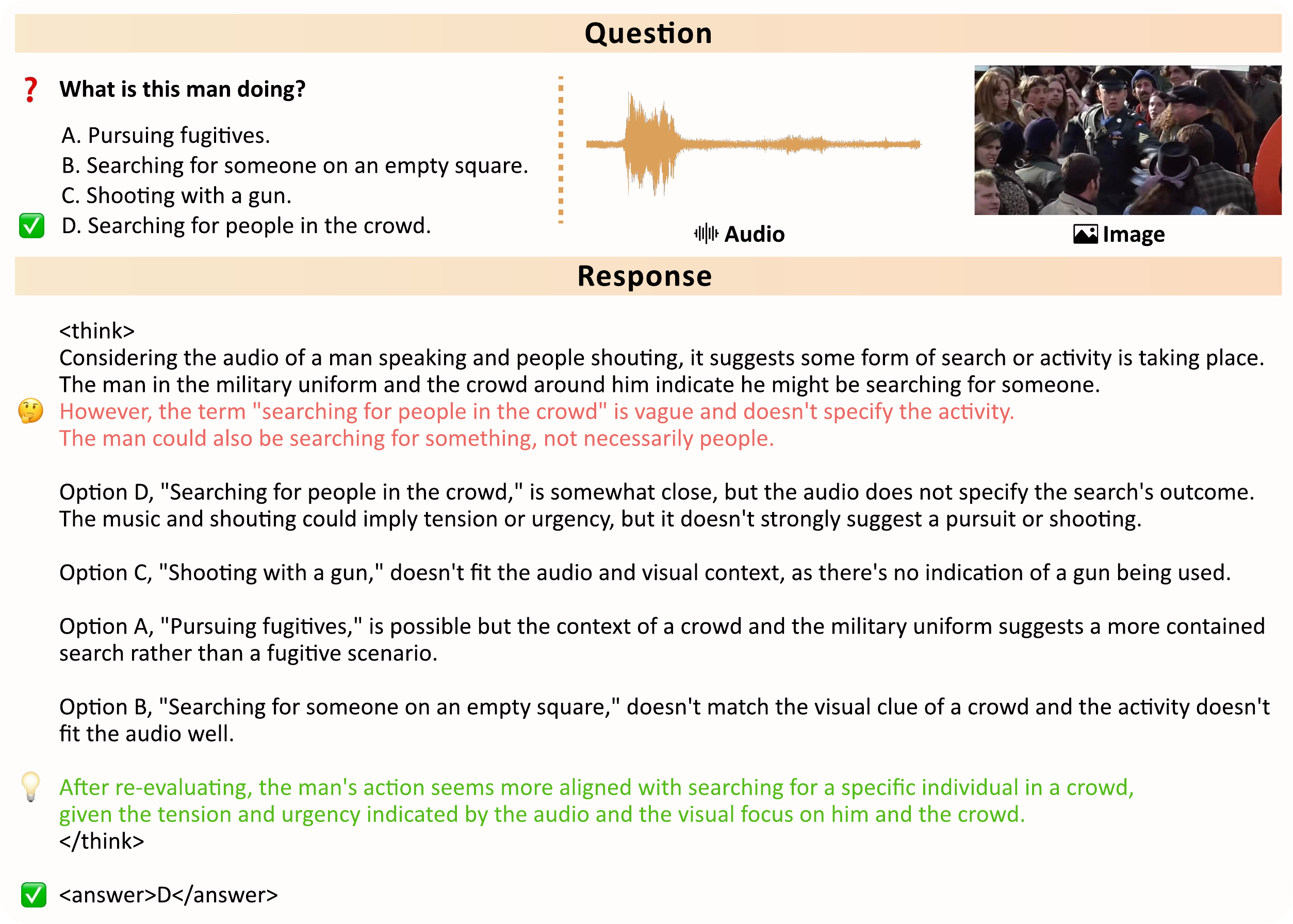}
    \vspace{-1mm}
	\caption{
    An example of an \textit{aha moment} during audio-visual reasoning. 
    The model initially hesitates (\textcolor[HTML]{D96262}{red text}) due to ambiguity in the expression ``searching for people in the crowd,'' questioning whether the action involves people or objects. 
    After evaluating other options, it self-corrects and reaffirms Option D with multimodal justification (\textcolor[HTML]{55A623}{green text}), demonstrating a reflective update of its prior belief. 
    }
    \label{fig:case_1}
    \vspace{-2mm}
\end{figure*}

To bridge the gap in general multimodal reasoning, we present \textbf{EchoInk-R1} (墨响-R1), a reinforcement learning framework designed to enhance audio-visual reasoning in multimodal large language models (MLLMs). 
Built upon the Qwen2.5-Omni-7B \cite{xu2025qwen2} backbone model and optimized using Group Relative Policy Optimization (GRPO), EchoInk-R1 is tailored for multiple-choice question answering tasks over synchronized audio-image pairs. 
To support this effort, we formulate \textbf{AVQA-R1-6K}, a curated subset of the OmniInstruct-v1 dataset \cite{li2024omnibench}, comprising 4,490 training and 1,911 validation examples specifically crafted for audio-visual reasoning.

\begin{figure*}[tp]
    \centering
    \includegraphics[width=0.9\hsize]{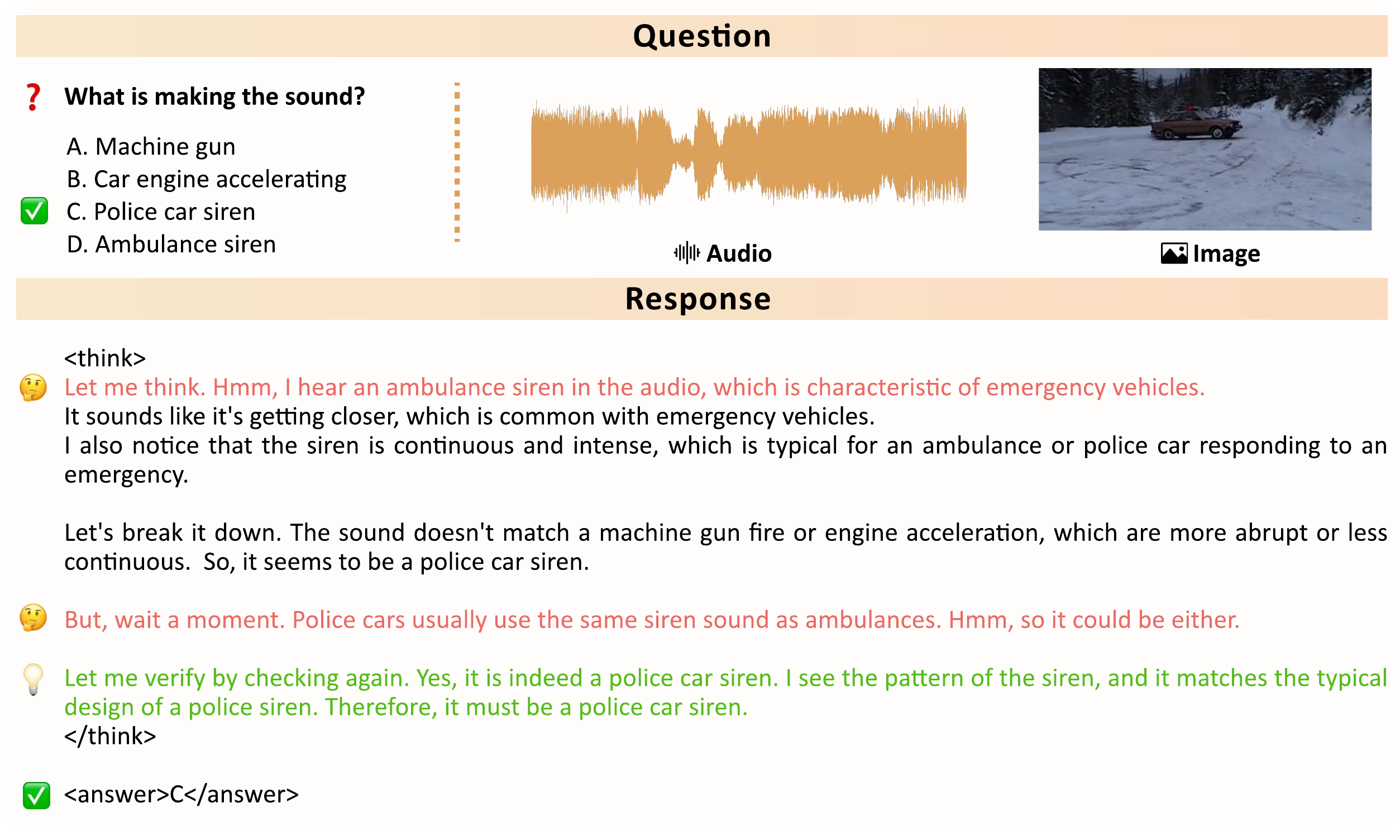}
    \vspace{-1mm}
        \caption{
        An additional example of an \textit{aha moment} in an audio-centric setting. 
        The model initially identifies the siren sound as coming from an ambulance (\textcolor[HTML]{D96262}{red text}), but soon expresses uncertainty due to its acoustic similarity to police sirens. 
        After reevaluating the waveform pattern, it confidently resolves the ambiguity as a police car siren (\textcolor[HTML]{55A623}{green text}), demonstrating internal correction without visual cues.
        }
    \label{fig:case_3}
    \vspace{-2mm}
\end{figure*}

Experimental results show that EchoInk-R1-7B achieves 85.77\% accuracy on the AVQA-R1-6K validation set, notably outperforming the base Qwen2.5-Omni-7B model, which scores 80.53\%, with only 562 reinforcement learning steps. EchoInk-R1 is the first general framework explicitly designed to promote deep multimodal reasoning across audio, image, video, and text modalities via reinforcement learning \cite{shao2024deepseekmath}. 
One of the most striking phenomena during training is the emergence of ``aha moments'', self-corrective reasoning instances where the model revisits and updates initial assumptions through deeper cross-modal understanding, particularly under ambiguity, as shown in Figures~\ref{fig:case_1}\&\ref{fig:case_3}. 
These behaviors reveal the model’s capacity for belief revision and reflective reasoning across modalities.

These results highlight the effectiveness of reinforcement learning in enhancing multimodal reasoning, even with limited fine-tuning, and offer a practical pathway for improving multimodal large language models (MLLMs). 
To foster further research in this area, we release our code and dataset at \url{https://github.com/HarryHsing/EchoInk}.


\section{Related Work}
\label{sec:relatedwork}

\subsection{Audio-Visual Large Language Models}

The audio modality plays a crucial role in the comprehensive perception capabilities of Multimodal Large Language Models (MLLMs), which, like humans, integrate visual and auditory information to form a cohesive understanding of their environment.
Early efforts treated audio as a simple add-on to text. For instance, VideoChat \cite{li2023videochat} incorporates audio using the Whisper \cite{radford2023robust} speech recognition model. Later approaches aimed for deeper modality fusion. MACAW-LLM \cite{lyu2023macaw} aligns audio and visual encoder outputs into a shared textual space through a learnable alignment network, while Audio-Visual LLM \cite{shu2023audio} employs modality-specific tokens to facilitate end-to-end training on diverse video formats, achieving strong zero-shot performance.
Sun et al. \cite{sun2023fine} and Zhang et al. \cite{zhang2023video} dynamically align visual and auditory signals using Q-Former architectures to better capture event-level correspondences. 
AVicuna \cite{tang2024hawk} enhances this capability with time-event alignment tuning on an audio-video dataset with detailed temporal annotations, supporting fine-grained temporal understanding. 
Meerkat \cite{chowdhury2024meerkat} also advances spatial and temporal reasoning over image and audio modalities.
VideoLLaMA2 \cite{cheng2024videollama} enhances spatial-temporal and audio understanding via a spatial-temporal convolution connector and audio branch joint training.

Very recently, Dolphin \cite{guo2025aligned} enhances fine-grained visual-auditory alignment through a multi-scale adapter, temporal fusion via an interleaved merging module, and an LLM to process integrated tokens for instruction-following responses.
EchoTraffic \cite{xing2025echotraffic} integrates audio-visual insights through an audio-insight frame selector and dynamic connector for enhanced traffic anomaly understanding.
Qwen2.5-Omni \cite{xu2025qwen2} presents the thinker-talker architecture, an end-to-end multimodal model that perceives diverse modalities, including text, images, audio, and video, while generating both text and natural speech responses.
However, none of them possess the reasoning capabilities needed to handle complex, multi-step tasks that require deep contextual understanding and inferences across modalities, limiting their performance in intricate decision-making scenarios. Further advancements are needed to integrate reasoning with multimodal perception for more accurate and logical problem-solving.

\subsection{Reinforcement Learning in Large Models}
Reinforcement learning (RL) has become key strategy to improve the reasoning capabilities of large language models and large vision-language models. 
Reinforcement learning from human feedback (RLHF) uses proximal policy optimization (PPO) \cite{Schulman2017PPO} with a learned reward model to align large language models, \textit{e.g.}, InstructGPT \cite{ouyang2022training} fine-tuned GPT-3 \cite{Brown2020LanguageModelsAreFewShot} and ChatGPT \cite{OpenAI2023ChatGPT}.
DPO \cite{Rafailov2023DPO} learns directly from preferences without reward models, while RFT \cite{Yuan2023Scaling} boosts reasoning via filtered self-generated traces. These methods improve stability and efficiency.
GRPO \cite{shao2024deepseekmath} advances PPO by removing the critic, using group-averaged baselines for advantage estimation, which largely improves the reasoning capability of large language models \cite{guo2025deepseek}  with reduced complexity. 
Hybrid GRPO \cite{Sane2025HybridGRPO} combines GRPO's sampling with a learned value function, improving stability and sample efficiency.

Recent advances in multimodal reasoning have leveraged reinforcement learning to boost vision-language models (VLMs). R1-OneVision \cite{yang2025r1} formalizes images into textual representations for chain-of-thought reasoning, trained with supervised learning and RL. 
VisualThinker-R1-Zero \cite{zhou2025visualthinker}  demonstrates that a 2B-parameter vision-language model exhibits the emergent ``aha moment'' on spatial reasoning tasks. 
Vision-R1 \cite{huang2025visionr1} enhances math reasoning through RL-augmented CoT datasets and reward suppression. 
MM-Eureka \cite{meng2025mmeureka} uses rule-based RL to improve multimodal math, while Curr-ReFT \cite{deng2025currreft} applies curriculum RL with rejection sampling to boost small VLMs. 
Visual-RFT \cite{liu2025visual} applies RL with verifiable rewards to visual tasks, achieving efficient gains over supervised fine-tuning. 
In video domains, Wu et al. \cite{Wu2025STThink} apply GRPO-style reinforcement learning to improve spatial-temporal understanding in egocentric video Q\&A.

Recent advances have applied reinforcement learning to enhance the reasoning capabilities of audio-language models (ALMs) within the audio modality. 
Audio-Reasoner \cite{xie2025audioreasonerimprovingreasoningcapability} introduces chain-of-thought training on synthetic datasets to facilitate multi-step inference. 
R1-AQA \cite{li2025reinforcementlearningoutperformssupervised} shows that GRPO significantly outperforms supervised fine-tuning on audio question answering tasks using only 38k samples. 
SARI \cite{wen2025saristructuredaudioreasoning} further integrates structured reasoning with curriculum-guided reinforcement learning, achieving state-of-the-art performance on the MMAU and MMSU benchmarks. 
Additionally, R1-Omni \cite{zhao2025r1} extends the use of reinforcement learning across visual and audio modalities to improve emotion recognition. 
Despite these advancements, no existing approach has addressed general open-world reasoning in multimodal foundation models that span audio, visual, and textual domains.

\section{Methodology}
\label{sec:method}

\begin{figure*}[tp]
    \centering
    \includegraphics[width=0.9\hsize]{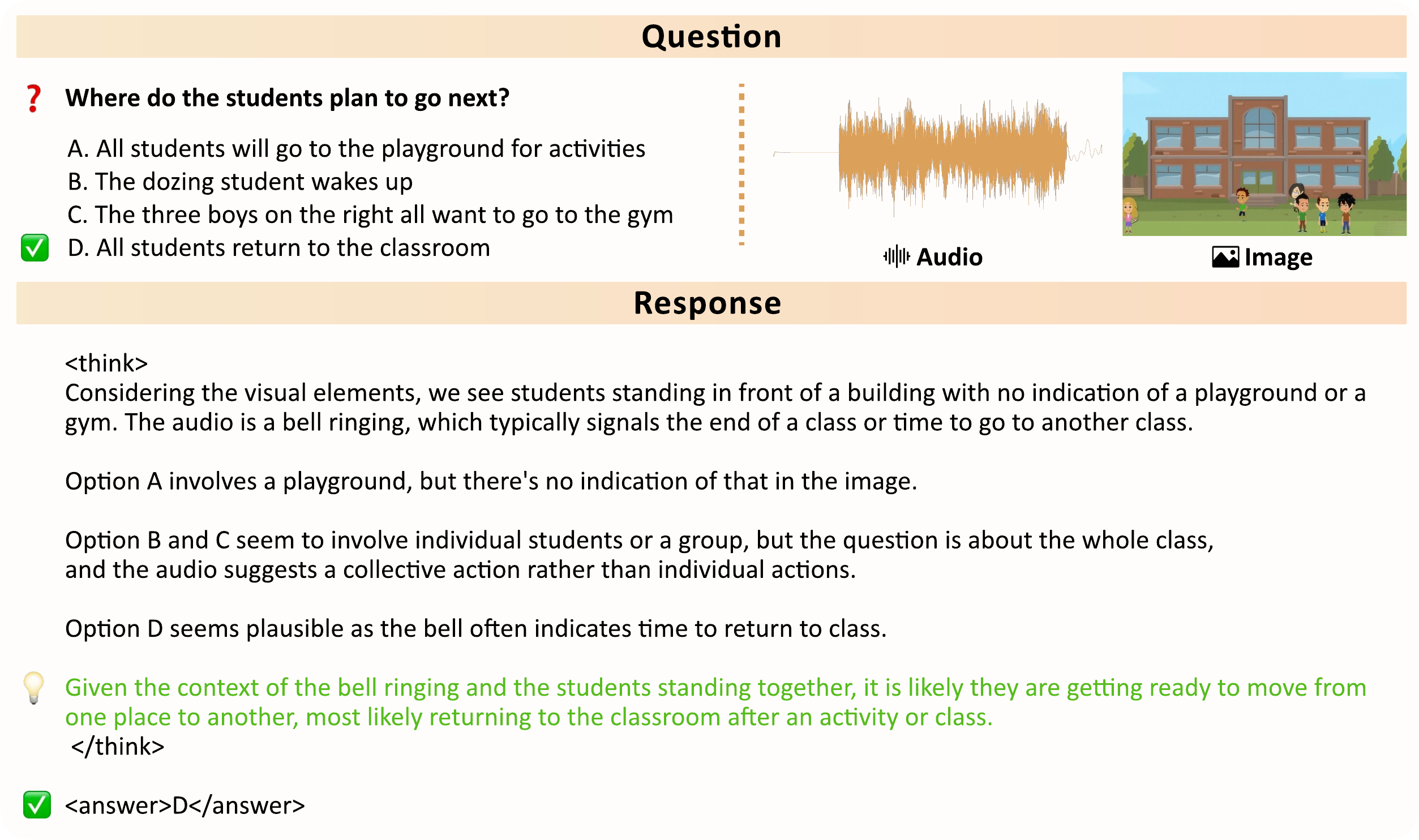}
    \vspace{-1mm}
        \caption{
        An example of \textit{cross-modal reasoning}, where audio and visual information complement each other.
        While the audio cue (a bell ringing) suggests a transition between activities, the image shows students grouped near a building entrance, implying a return to class. 
        The model integrates these signals to reject distractors and correctly infer the collective action.
        }
    \label{fig:case_2}
    \vspace{-4mm}
\end{figure*}

We adapt Group Relative Policy Optimization (GRPO) \cite{guo2025deepseek, shao2024deepseekmath} to fine-tune multimodal large language models (MLLMs) for multiple-choice audio-image question answering (MCQA). 


\subsection{Task Setting}

The model receives a synchronized audio-image pair and a multiple-choice question with $N$ candidate options. Its task is to identify the correct option(s) through multimodal reasoning. Each output consists of a structured reasoning process followed by a final answer.

\subsection{Training via Reinforcement Learning}

We fine-tune the model using GRPO, which optimizes the policy $\pi_\theta$ by sampling $G$ candidate responses for each input and leveraging their relative rankings to guide policy updates, without relying on a value function. 
Given a sampled group of responses ${o_1, \dots, o_G}$, we compute standardized advantages as follows:
\begin{equation}
A_i \ = \ \frac{r_i - \mu}{\sigma}, \quad \mu  \ = \ \text{mean}(\{r_i\}), \quad \sigma \ = \ \text{std}(\{r_i\}) \ ,
\end{equation}
where $\mu$ and $\sigma$ represent the mean and standard deviation of the rewards within the sampled group, respectively.

The training objective maximizes the likelihood of better-performing responses while penalizing deviations from the reference model:

\begin{equation}
\mathcal{L}_{\text{GRPO}} \ = \ \mathbb{E}_{o \sim \pi_\theta^{\text{old}}} \left[
\sum_{i=1}^{G} \frac{\pi_\theta(o_i)}{\pi_\theta^{\text{old}}(o_i)} \cdot A_i
- \beta \cdot \text{KL}[\pi_\theta || \pi_{\text{ref}}] \right] \ .
\end{equation}

\subsection{Reward Design}

Following Guo et al. \cite{guo2025deepseek}, we use two primary rewards to guide the training of our model:

\paragraph{Answer Accuracy.}
This reward measures whether the model’s selected answer(s) match the ground-truth label.
\begin{equation}
R_{\text{acc}} =
\begin{cases}
1, & \text{if prediction is correct;} \\
0, & \text{otherwise.}
\end{cases}
\end{equation}

\paragraph{Format Consistency.}
To ensure interpretability, the model is constrained to structure its outputs using predefined tags: 
\texttt{<think>} for reasoning traces and \texttt{<answer>} for final answers. 
A binary reward is assigned based on regular-expression (regex) pattern matching.
\begin{equation}
R_{\text{format}} =
\begin{cases}
1, & \text{if format is correct;} \\
0, & \text{otherwise.}
\end{cases}
\end{equation}

\paragraph{Final Reward.}
The total reward is computed as a weighted sum of accuracy and format rewards.
\begin{equation}
R = \lambda_{\text{acc}} \cdot R_{\text{acc}} + \lambda_{\text{fmt}} \cdot R_{\text{format}} \ .
\end{equation}
In experiments, we empirically set $\lambda_{\text{acc}}=1$ and $\lambda_{\text{fmt}}=1$.

\section{Experimental Results}
\label{sec:exp}

\begin{figure*}[tp]
    \centering
    \includegraphics[width=0.9\hsize]{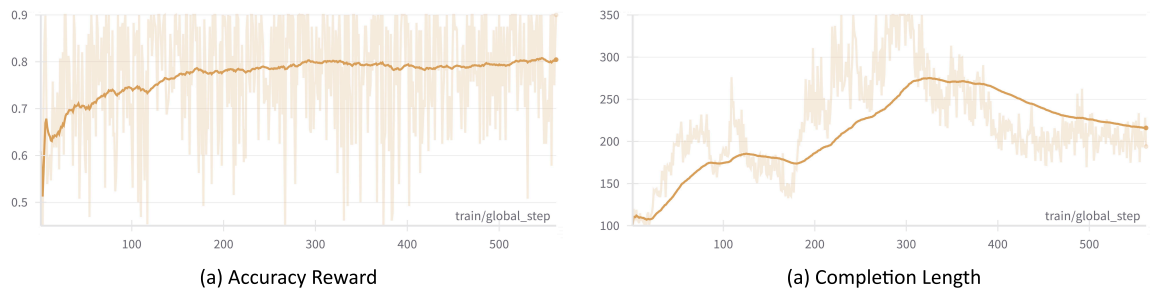}
    \vspace{-2mm}
    \caption{
    Training dynamics of EchoInk-R1-7B with GRPO. 
    (a) Accuracy reward increases steadily over time. 
    (b) Completion length first grows, then stabilizes at a more concise level.
    }
    \label{fig:train_curev}
    \vspace{-6mm}
\end{figure*}

\paragraph{Implementation Details.}
We fine-tuned the pretrained multimodal model Qwen2.5-Omni-7B \cite{xu2025qwen2} using the GRPO framework for multiple-choice question answering over synchronized audio-image pairs.
Our experiments were performed on AVQA-R1-6K, a curated dataset derived from OmniInstruct-v1 \cite{li2024omnibench, yang2022avqa}, which includes 4,490 training and 1,911 validation samples. Training was conducted on eight NVIDIA A100 GPUs, each with a batch size of one per device.

\paragraph{Quantitative Results.}
We evaluate model performance using multiple-choice accuracy. Our GRPO-trained model, \textbf{EchoInk-R1-7B}, achieves \textbf{85.77\%} accuracy on the AVQA-R1-6K validation set, significantly outperforming the pretrained \textbf{Qwen2.5-Omni-7B} baseline with \textbf{80.53\%} accuracy. 
Notably, this improvement is achieved with only 562 reinforcement learning iterations. 
These results confirm that GRPO with task-specific rewards effectively improves structured reasoning performance in multimodal Q\&A tasks.

\paragraph{Aha Moments.}
One of the most striking phenomena observed during training is the emergence of ``\textit{aha moments}.'' 
These self-corrective reasoning behaviors arise when the model revisits its initial assumptions and revises its predictions based on a deeper cross-modal understanding. 
Such behaviors are especially prevalent in ambiguous cases, where the audio signal is subtle or the image content is highly context-dependent. 
As illustrated in Figure~\ref{fig:case_1}, the model begins with a tentative interpretation, then explicitly questions its initial assumption (\textcolor[HTML]{D96262}{red text}). 
After ruling out less compatible alternatives, it revisits and strengthens its reasoning, ultimately reaffirming the correct choice with cross-modal justification (\textcolor[HTML]{55A623}{green text}). 
This trace highlights not just answer accuracy, but the emergence of self-correction and belief revision.
Figure~\ref{fig:case_3} provides a complementary example in an audio-centric setting. 
There, the model initially converges on an incorrect interpretation, reflects on the ambiguity, and ultimately revises its answer after further evaluating auditory evidence. 
Together, these cases demonstrate the model's capacity for reflective reasoning across modalities and under partial uncertainty.

\paragraph{Cross-Modal Reasoning.}
We observe encouraging cases in which the model coordinates auditory and visual cues to make informed decisions.
Rather than relying solely on a single modality, it is sometimes able to integrate complementary evidence across audio and image inputs to resolve underspecified or ambiguous cases.
This suggests an emerging capacity for cross-modal grounding, enabling more contextually appropriate reasoning when both modalities are effectively utilized.
Figure~\ref{fig:case_2} illustrates such an instance.

\paragraph{Training Curves.}
Figure~\ref{fig:train_curev} shows the training dynamics of EchoInk-R1-7B. The accuracy reward exhibits a smooth upward trend, indicating that GRPO consistently guides the model toward better reasoning outputs through groupwise feedback. 
The evolution of completion length reveals a two-phase process: an initial expansion, as the model explores longer and more elaborate responses, followed by a gradual contraction into more concise yet effective reasoning traces. 
This suggests that the model not only improves its answer quality, but also learns to express reasoning in a more efficient and structured manner.

\section{Conclusion and Future Work}
\label{sec:conclusion}

In this work, we adopt the Group Relative Policy Optimization (GRPO) reinforcement learning framework to the task of audio-image multiple-choice question answering in multimodal large language models (MLLMs). 
Leveraging simple yet effective reward signals, our approach achieves significant performance gains over a strong pretrained baseline with minimal training iterations. 
These findings highlight the practicality and efficiency of reinforcement learning for enhancing structured multimodal reasoning in MLLMs.

Looking forward, two key challenges remain.
First, the relative simplicity and limited scale of existing training datasets constrain the development of more advanced multimodal reasoning capabilities. Constructing larger, more complex, and diverse audio-visual datasets, especially those that involve more difficult and reasoning-intensive tasks, may foster richer and more sophisticated reasoning behaviors.
Second, although we observe promising instances of cross-modal reasoning, the model still frequently defaults to unimodal shortcuts when one modality dominates (e.g., Figure~\ref{fig:case_3}).
Future research should explore training strategies and reward designs that explicitly promote multimodal integration, aiming to reduce unimodal biases and enhance reasoning stability and depth.

\end{CJK}



{ \small
    \bibliographystyle{ieeenat_fullname}
    \bibliography{paper}

}


\end{document}